\documentclass[runningheads]{llncs}

\usepackage{amssymb}
\usepackage{graphicx}

\usepackage{url}
\urldef{\mailsa}\path|zhengbojin@gmail.com|
\urldef{\mailsb}\path|yxli@whu.edu.cn|

\newcommand{\keywords}[1]{\par\addvspace\baselineskip
\noindent\keywordname\enspace\ignorespaces#1}

\begin{document}

\mainmatter  % start of an individual contribution

\title{New Model for Multi-Objective Evolutionary Algorithms}

\titlerunning{New Model for MOEA}

\author{Bojin Zheng \inst{1} \and Yuanxiang Li\inst{2} }
\authorrunning{Bojin Zheng et al.}
\institute{ College of Computer Science, South-central University for Nationalities,Wuhan,430074,China \and
State Key Lab. of Software Engineering,Wuhan University,Wuhan,430072,China\\
\mailsa, \mailsb\\
}

\toctitle{Lecture Notes in Computer Science} \tocauthor{Bojin Zheng (South-Central University For Nationalities,
China),Yuanxiang Li (Wuhan University, China)}

\maketitle

\begin{abstract} Multi-Objective Evolutionary Algorithms (MOEAs) have been proved efficient to deal with Multi-objective Optimization
Problems (MOPs). Until now tens of MOEAs have been proposed. The unified mode would provide a more systematic approach
to build new MOEAs. Here a new model is proposed which includes two sub-models based on two classes of different
schemas of MOEAs. According to the new model, some representatives algorithms are decomposed and some interesting
issues are discussed.
\end{abstract}

\section{Introduction}
\par Evolutionary Algorithms are an randomized searching approach based on
Darwin's evolutionary theory. They play an important role in many
fields such as optimization, control, game strategies, machine
learning, and engineering design etc.
\par In 1984, David Schaffer introduced Vector Evaluated Genetic Algorithm (VEGA)\cite{1749,1750} to solve Multi-objective
Optimization Problems(MOPs). Henceforth, the research on Multi-Objective Evolutionary Algorithms(MOEAs) attracted more
and more researchers. Up to now tens of MOEAs have been proposed.

To guide the efforts on MOEAs, some researchers tried to build unified models for popular MOEAs. For examples, Macro
Laumanns et al.\cite{1752} proposed a unified model for the Pareto-based and elitist MOEAs in 2000. This model can
describe most popular famous MOEAs, such as Non-dominated Sorting Genetic Algorithm II(NSGA-II\cite{549}, Strength
Pareto Evolutionary Algorithm and its improvement(SPEA/SPEA2)\cite{587}, Pareto Archived Evolution
Strategy(PAES)\cite{1751} and so on. \cite{519} expressed the schema of MOEAs which employ archive with such a formula as follows: \\
\centerline{ MOEA = Archive + Generator} \\But this formula is quite simple. Recently, more and more MOEAs can not be
accurately described by these models, such as Adaptive Grid Algorithm (AGA)\cite{563},Rank-Density based Genetic
Algorithm (RDGA)\cite{558}, Geometrical Pareto Selection (GPS)\cite{1753,4028} and GUIDED \cite{4056} etc.

In this paper, we propose a new model to describe the advanced MOEAs. In section 2,the model is introduced . And then
SPEA\cite{587} ,AGA\cite{563} and GPS \cite{1753,4028} are decomposed in section 3 according to this model. Some
interesting issues are discussed in section 4. In section 5, some conclusions are made.
\section{Introduction to the New Model}
The first MOEA -- VEGA -- is a non-Pareto algorithm. Subsequently, Goldberg D. E. \cite{2406} proposed to use Pareto
dominance to compute the fitness of the individuals based on 'Ranking' method. Subsequent experiments prove that  Pareto
dominance based MOEAs are more efficient than non-Pareto MOEAs. Since the work of Zitzler et al.\cite{587}, the 'elitism' of MOEAs
has been recognized: Elitism of MOEAs is especially beneficial in deal with MOPs and the use of elitism can speed up the
convergence to the Pareto front.
the Pareto based MOEAs with elitism is more efficient than the MOEAs without elitism. To implement
the elitism, many MOEAs use a secondary 'elitist' population ,i.e., the archive, to store the elite individuals.
According to MOEAs' formula, the pseudocode of common elitist MOEAs with archive can be depicted as Figure
\ref{fig.pseudocode}:
\begin{center}
\begin{figure}[!htbp]
\textbf{\noindent 1 initialize the population and archive \\
2 evaluate the population \\
3 while the termination criterions have not been reached do \\
4 \ \ \ \ generate a solution by the generator \\
5 \ \ \ \ evaluate the new solution            \\
6 \ \ \ \ try to update the archive\\
7 \ \ \ \ according to the feedback of archive, try to update the population \\
8 end while }
 \caption{Pseudocode for Generic MOEAs with Archive} \label{fig.pseudocode}
\end{figure}
\end{center}

The pseudocode seemly does not mention the generation gap methods. Actually, the generation gap methods can be
decomposed into this model, if we see the generation number and the replaced parent individuals as the additional
parameters. Moreover, though Single-Objective Evolutionary Algorithms(SOEAs) and MOEAs are very similar, there still
are three major differences:
\begin{enumerate}
  \item Different to single-objective optimization, the generator of MOEAs may crossover some
individuals in population with the individuals in population or archive
  \item the fitness assignment is more
complicated, because it is relative to two operators: fitness evaluation for the archive and fitness evaluation for the
population
  \item As to the elitism,  the SOEAs just keeps only one fittest individual. But in MOEAs, the elitism, commonly, the strategy to update
  the archive is quiet complicated.
\end{enumerate}

\begin{figure}[!htbp]
\centering{\includegraphics [width=3.864in,height=1.75in]{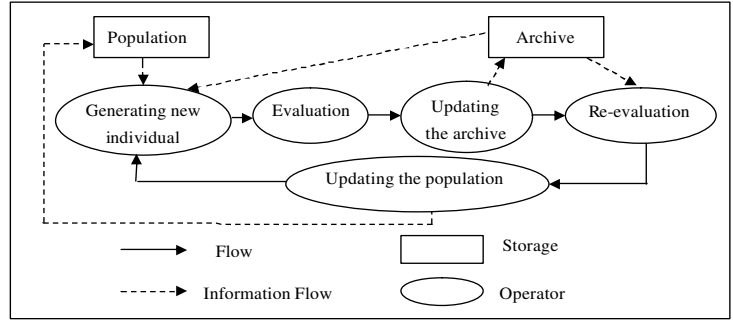} \caption{The Model of Elitism MOEAs with
Archive} \label{fig.framework} }
\end{figure}

In general, the new model can be depicted as Figure \ref{fig.framework}:\\

\begin{figure}[!htbp]
\centering{\includegraphics [width=10.7424cm,height=4.5576cm]{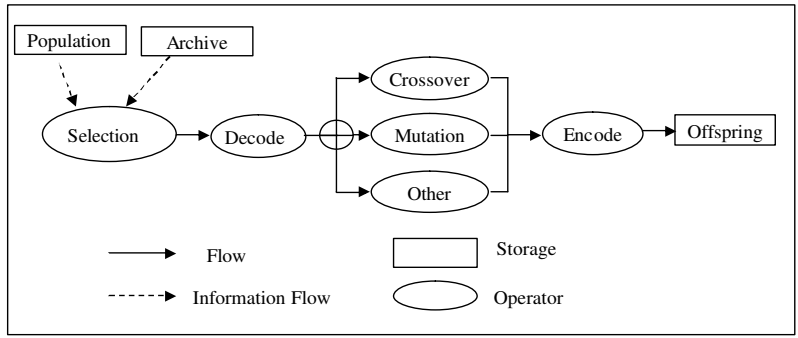} \caption{The Framework for the Generator}
\label{fig.NewIndi} }
\end{figure}

In Figure \ref{fig.framework}, updating the archive would retrieve information from the archive, so the link is not
drawn in this framework. Moreover, the generator is similar to the generating process of SOEA. Actually, except the
selection operators, they both are same. It can be depicted as Figure \ref{fig.NewIndi}.

Secondly, the strategy of updating the archive is different to SOEA and very complicated.

\begin{figure}[htbp]
\centering{\includegraphics [width=3.504in,height=1.842in]{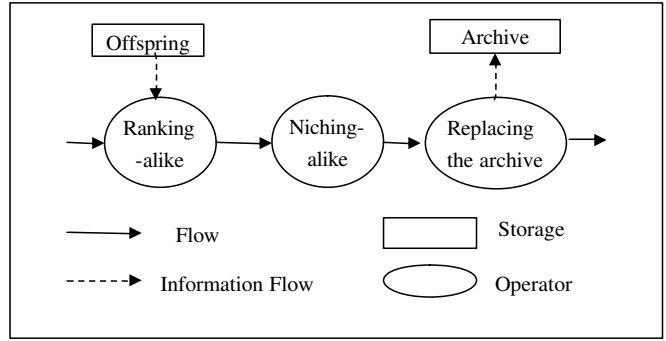} \caption{The Ranking-alike and Niching-alike
Schema} \label{fig.rankniche} }
\end{figure}

Very many MOEAs employ the ranking-alike operators and the niching-alike operators. In such a schema, ranking-alike
operators are firstly employed to eliminate the dominated solutions and secondly niching-alike operators are employed
to eliminate the crowded solutions. But unfortunately, this kind of MOEAs are not convergent\cite{2479} because of
fitness deterioration. The schema of ranking-alike and niching-alike MOEAs(RN\_MOEA) could be depicted as Figure
\ref{fig.rankniche}.

\begin{figure}[htbp]
\centering{\includegraphics [width=3.704in,height=1.576in]{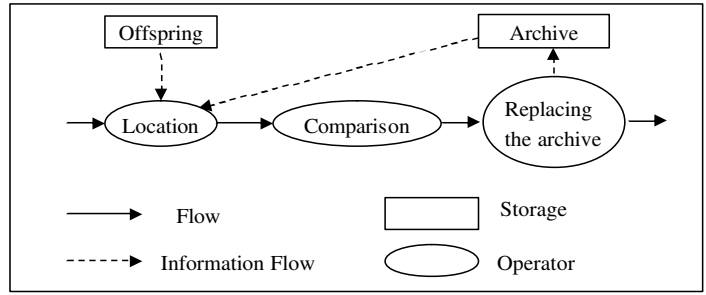} \caption{The Sampling Schema}
\label{fig.sampling} }
\end{figure}

Actually, except this schema, there exists another schema. We call it 'the sampling schema'. In this schema, the
feasible solution space (includes Pareto optimal front) is divided into grids in advance, when new solution is
generated and evaluated, the algorithm firstly computes its coordinate in the grids and compare it with the
individual(s) in the right coordinate. Whether updating the archive with the new solution or not just depends on the
comparison. Obviously, this kind of methods are very different to ranking-alike and niching-alike methods, they use
'local dominance' instead of 'global dominance', and therefore hold lower time complexity. Some of them do not
eliminate the dominated solutions from the archive in the main loop of algorithm, so additional operation(eliminating
operator) should be employed after the main loop to cut the dominated solutions off from the archive. But if the
archive should only store nondominated solutions, the eliminating operator should be integrated into the main loop. As
to the diversity of Pareto optimal front, it depends only on the generator, because the span of cells in the grids has
been predefined, may adaptively. The schema of sampling MOEAs(SA\_MOEA) can be depicted as Figure \ref{fig.sampling}.

\section{Decomposing Existing MOEAs}
In this section, we will use SPEA , AGA and GPS to show how to decompose the existing algorithms.
\subsection{SPEA}
As to SPEA, it is a classical RN\_MOEA. SPEA's mating selection process is a typical selection operator which uses
archive and population. clustering method actually is an operator to keep diversity,i.e., niching-alike operator.
Moreover, the truncate and update function can be seen as the strategies to update the archive and population.
\subsection{AGA}
Actually, AGA\cite{563} is a typical archiving algorithm. It must combine a generator to become MOEAs. AGA employ a
nondominated vectors archive, so it employs \textbf{Is\_Dominated} function and \textbf{Dominates} function to
eliminate the dominated solutions. Moreover, AGA uses adaptive grids, that is, the boundaries should be extended or
reduced. However, in spite of these details, AGA samples the feasible solution space in essence.
\textbf{Reduce\_Crowding} function, \textbf{Steady\_State} function and \textbf{Fill} function actually perform the
comparison and replacement.
\subsection{GPS}
GPS is also an archiving algorithm. It employs the \textbf{Location} operator to retrieve the right solutions which
will be used to compare with the new solution, employs \textbf{Comparison} operator to perform the comparison, and at
last employs the \textbf{Steady\_State} operator to update the archive with the new solution if the new solution is
better than the original solution in the archive. Moreover, the archive of GPS may store the dominated solution, only
when the main loop ends, an additional operation is used to eliminate the dominated solutions.

\section{Some Important Issues}
According to the model, there are two kinds of archiving algorithms. Based on different schema, MOEAs would behave
differently, and they would have different properties.

\subsection{Performance Measure}
The convergence property is very important to MOEAs. It would theoretically determine the approximation degree to the true Pareto front.
But good diversity would be very helpful of the decision-maker. That is, MOEAs had better converge with diversity. So the performance measure
should take both the convergence and diversity into considerations.

AGA has been proved convergent with well-distributed solutions under certain strict conditions. GPS also converges to
Ray-Pareto optimal front. In contrast to SA\_MOEAs, the RN\_MOEAs do not converge. Furthermore, the SA\_MOEAs could be
improved to converge to true Pareto front under certain conditions.

The complexity of MOEAs would be another aspect of performance measure. If two multi-objective approaches have different
archivers, their average performance may differ\cite{519}. Because of ranking method, the time complexity of RN\_MOEAs would be greater than or equal to
$O(MN)$ where computing one new individual, here M is the number of objectives, N is the size of population(or archive).
As to AGA and GPS, the time complexity is $O(M)$. Niching-alike operators often hold a space complexity of $O(N^2)$. But AGA's is $O(N^M)$, GPS's is $O(N)$
at the best situation,$O(N^{M-1})$ at the worst situation. Furthermore, we can reduce GPS's space complexity to $O(N)$ by
using binary tree with an additional average
time complexity of $O(Nlog_{2}N)$.

\subsection{Cooperation between Generator and Archive}
In this model, the generator should cooperate with archive to control the evolving directions. Therefore, considering
to deal with difficult objective functions, 'local search' may be used to exploit.  As to the archive of SA\_MOEAs, 'local
dominance' is useful to reduce the time complexity. 'local search' and 'local dominance' are different concepts.

The evolving directions of the population are multi-objective. In one hand, the selection pressure should make the
individuals evolving toward the true Pareto front, i.e., depth-first search. In the other hand, the selection pressure
should make the individuals spread over the whole Pareto front, i.e., width-first search. How to deal with the conflict
between depth-first search and width-first search is still lack of delicate research. As to SA\_MOEAs, because of local dominance,
the feedback of archive just provide information for depth-first search, less for width-first search.

\subsection{Taxonomy}
Based the proposed model, we suggest that MOEAs could be categorized into four classes:
\begin{enumerate}
  \item Non-Pareto MOEAs\\
The representative MOEA is VEGA\cite{1749,1750}. This algorithm employs multiple sub-populations to optimize every
single objective separately. This algorithm often converge to special points which often are not Pareto Optimal points,
moreover, the diversity is not taken into consideration.
  \item Pareto MOEAs (without Elitism)\\
  The representative MOEAs include Multi-Objective Genetic Algorithm (MOGA) \cite{3063},  Niched Pareto Genetic
    Algorithm (NPGA) \cite{2519,2410} and Non-dominated Sorting Genetic Algorithm (NSGA)\cite{3440}. These algorithms
    employ some strategies to maintain the diversity, but approximation is not good enough.
    \item Pareto MOEAs with Elitism \\
The representative MOEAs include NSGA-II\cite{549} , Strength Pareto Evolutionary Algorithm and its
improvement(SPEA/SPEA2)\cite{587}. These algorithms employ elitism strategy to maintain good approximation. But these
algorithms are not convergent.
  \item Convergent MOEAs\\
Actually, the archiving algorithms determine the convergence property of MOEAs. The representative algorithms
  include Adaptive Grid Algorithm (AGA) \cite{563}, GPS \cite{1753,4028}. These algorithms should converge/pseudoconverge under certain conditions.
\end{enumerate}

\section{Conclusions and Future Work}
The proposed model is intended to understand the-state-of-the-art MOEAs and provides a more systematic approach to design more efficient and more customized
MOEAs for researchers and possible users.

Our model implies that the ranking-alike and niching-alike schema is very different to the sampling schema,though they both may use archive
to store elitist solutions.

In contrast to the previous unified models, the new model can describe the-state-of-the-art MOEAs more accurately and be more atomic. So it is more
convenient to use this model for the analysis of the algorithms.

This model provide us many cues to improve the MOEAs, such as the relationship between 'local search' and 'local dominance', the relationship between
evaluation operation and re-evaluation operator and the relationship between depth-first search and width-search etc.  The future work would try to
discover more principles and develop new operators based on this model.
\newline
\par
\noindent \textbf{Acknowledgement.} \par \noindent The authors gratefully acknowledge the financial support of the
National Natural Science Foundation of China under Grant No.60473014 and No.60603008.
\bibliography{Model}

\end{document}